\documentclass[
    letter, 11pt,
    colorlinks,
    linkcolor=blue,
    citecolor=blue,
    pagebackref
    ]{article}

\usepackage[utf8]{inputenc}
\usepackage[margin=.95in]{geometry}
\usepackage[numbers, sort]{natbib}
\usepackage{authblk}
\makeatletter
\def\blfootnote{\xdef\@thefnmark{}\@footnotetext}
\makeatother
\usepackage{amsmath,amsfonts,amsthm,amssymb}
\usepackage{mathtools,cancel}
\usepackage{thmtools, amssymb}
\usepackage{bbm}
\usepackage{algorithm}
\usepackage{xcolor}
\usepackage[noend]{algpseudocode}
\usepackage{mdframed}
\usepackage{caption}
\usepackage{xspace}

\usepackage[colorlinks,allcolors=blue]{hyperref}
\usepackage{enumitem}
\newcommand{\labelthis}[1]{\stepcounter{equation}\tag{\theequation}\label{#1}}
\usepackage[capitalise]{cleveref}
\newcommand{\hist}{\mathrm{hist}}
\declaretheorem{theorem}

\newtheorem*{theorem*}{Theorem}
\newtheorem{proposition}{Proposition}
\newtheorem*{corollary*}{Corollary}
\declaretheorem[sibling=theorem]{lemma}

\declaretheoremstyle[
        spaceabove=\topsep, 
        spacebelow=\topsep, 
        bodyfont=\normalfont,
        notefont=\normalfont\bfseries,
        notebraces={}{},
        qed=$\blacksquare$, 
    ]{proofstyle}
\declaretheorem[style=proofstyle,numbered=no,name=Proof]{proof}

\usepackage{xcolor}
\usepackage[extdef=true]{delimset}

\usepackage{crossreftools}

\crefname{claim}{Claim}{Claims}
\newcommand{\safe}{\mathrm{safe}}

\newcommand{\E}{\mathop{\mathbb{E}}}

\newcommand{\X}{\mathcal{X}}

\renewcommand{\P}{\mathbb{P}}

\newcommand{\ignore}[1]{}

\usepackage{wrapfig}
\setitemize[1]{leftmargin=2ex}
\setenumerate[1]{leftmargin=2ex}

\usepackage{wrapfig}
\usepackage{selectp}
\title{Can Copyright be Reduced to Privacy?}
\author[1]{Niva Elkin-Koren}
\author[1]{Uri Hacohen}
\author[2]{Roi Livni}
\author[3]{Shay Moran}
\affil[1]{Faculty of Law, Tel Aviv University}
\affil[2]{Department of Electrical Engineering, Tel Aviv University}
\affil[3]{Departments of Mathematics and Computer Science, Technion}

\usepackage{titlesec}

\titlespacing*{\section}{0pt}{0.2\baselineskip}{0.2\baselineskip}

\titlespacing*{\subsection}{0pt}{0.3\baselineskip}{0.2\baselineskip}
\titlespacing*{\paragraph}{0pt}{0.3\baselineskip}{0.2\baselineskip}
\begin{document}
\maketitle

\thispagestyle{empty}
\begin{abstract}
There is a growing concern that generative AI models will generate outputs  closely resembling the copyrighted materials for which they are trained.  This worry has intensified as the quality and complexity of generative models have immensely improved, and the availability of extensive datasets containing copyrighted material has expanded. Researchers are actively exploring strategies to mitigate the risk of generating infringing samples, with a recent line of work suggesting to employ techniques such as differential privacy and other forms of algorithmic stability to provide guarantees on the lack of infringing copying.
In this work, we examine whether such algorithmic stability techniques are suitable to ensure the responsible use of generative models without inadvertently violating copyright laws. We argue that while these techniques aim to verify the presence of identifiable information in datasets, thus being privacy-oriented, copyright law aims to promote the use of original works for the benefit of society as a whole, provided that no unlicensed use of protected expression occurred. These fundamental differences between privacy and copyright must not be overlooked. In particular, we demonstrate that while algorithmic stability may be perceived as a practical tool to detect copying, such copying does not necessarily constitute  copyright infringement. Therefore, if adopted as a standard for detecting an establishing copyright infringement, algorithmic stability may undermine the intended objectives of copyright law.
 
\end{abstract}

\clearpage
\pagenumbering{arabic} 

\section{Introduction}
Recent advancements in machine learning have sparked a wave of new possibilities and applications that could potentially transform various aspects of our daily lives and revolutionize numerous professions through automation. However, training such algorithms heavily relies on extensive content
which may include copyrighted materials. Under U.S copyright law, copyright protection subsists in original content of authorship fixed in any tangible medium of expression \cite{17_U.S.C.}, excluding any ``idea, procedure, process, system, method of operation, concept, principle, or discovery, regardless of the form in which it is described, explained, illustrated, or embodied in such work.'' \cite[§ 102(b)]{17_U.S.C.}. The unauthorized copying of copyrighted works may amount to copyright infringement \cite[§ 106]{17_U.S.C.} unless permitted by exceptions and limitations provided by law (\cite[§107-122]{17_U.S.C.}, and \cite{samuelson2023generative}). Consequently, identifying and, determining when and how content can be used within this framework without infringing upon individuals’ legal rights has become a pressing challenge. Foundation Models and generative AI (GenAI), trained on gigantic datasets, exacerbate this challenge. One area where this issue arises prominently is in the operation of generative models, which take human-produced content—much of it copyrighted as input and are expected to generate ”-similar” content. For instance, consider a machine trained on images and then generates new images that resemble the ones it was trained on. In this context, the fundamental question arises:
\begin{center}
   
\emph{When does the content generated by a machine (output content) infringe copyright in the training set (input content)?}

\end{center}

This question is not purely theoretical, as various aspects of this problem have  become subjects of legal disputes in recent years. In 2022, a class action was filed against Microsoft, GitHub, and OpenAI, claiming that their code-generating systems, Codex and Copilot, infringed copyright in the licensed code that the system was allegedly trained on \cite{DOE_GitHub}. Similarly, in another class action, against Stable Diffusion, Midjourney, and DeviantArt, plaintiffs argue that by training their system on web-scraped images, the defendant infringes millions of artists' rights \cite{Anderson_Stability_AI}. Allegedly, the images produced by these systems, in response to prompts provided by the systems’ users, are derived solely from the training images, which belong to plaintiffs, and, as such, are considered unauthorized derivative works of the plaintiffs’ images \cite[§ 106 (2)]{17_U.S.C.}.

A preliminary question is whether it is lawful to make use of copyrighted content in the course of training  \cite{lemley2020fair,grimmelmann2015copyright, usesofcopyrightedmaterials2022}. There are compelling arguments to suggest that such intermediary copying might be considered fair use \cite{lemley2020fair}.  For example, Google’s Book Search Project—entailing the mass digitization of copyrighted books from university library collections to create a searchable database of millions of books—was held by US courts to be fair use \cite{Authors_Guild_Google}. Then, there is a claim that generative models reproduce protected copyright expressions from the input content on which the model was trained. However,  to claim that the output of a generative model infringes her copyright, a plaintiff must prove not only that the model had access to her copyrighted work, but also that the alleged copy is substantially similar to her original work \cite{Sid_McDonald,Brown_Symantec}

Identifying what constitutes ``substantial similarity,” and unlawful copying remains a pressing challenge. Recent studies have proposed measurable metrics to quantify copyright infringement  \cite{barak,bousquet, scheffler2022formalizing,carlini2023extracting}. One approach, \cite{barak,bousquet} asserts that a machine generating output content substantially
similar to an input content does not infringe that input content copyright
if the machine would have reasonably generated the same output
content even without accessing the input content. This argument can be illustrated as follows: Suppose that Alice outputs content A and Bob claims it plagiarizes content B. Alice might argue that she never saw content B, and would reason that this means she did not infringe Bob’s copyright. However, since Alice must have observed some content, a second line of defense could be that “\textbf{had} she never saw B” she would still be likely to produce A.
The above argument was exemplified by \citet{bousquet} who interprets differential-privacy in the above manner.
Subsequently, \citet{barak} presented a certain generalization, in the form of a \emph{near-free access} (NAF) notion that can potentially allow a more versatile notion of copyright protection. Both applications draw on algorithmic stability notions used in privacy research.

However, certain crucial traits of copyright law make it challenging to reduce the problem to a question of privacy. An essential element of copyright law in the United States is utilitarian rationale, seeking to promote the creation and deployment of creative works \cite{US_constitution_1,mazen_stein}. It is crucial, then, that any interpretation of copyright, or for that matter any quantifiable measure for copyright, will be aligned with these objectives. In particular, while the law delineates exclusive rights to the creators of original expressions, it must ensure sufficient creative space for current and future creators  \cite{samuelson2016reconceptualizing}. For this reason, several criteria exist in copyright law, specifically allowing breathing room for subsequent authors to draw upon copyrighted content. These criteria distinguish copyright law from privacy as defined by algorithmic stability notions. First, copyright is limited in time, and once protection has expired, the copyright content enters the public domain and is free for all to use without authorization \cite{litman1990public}. 
This issue, though, can be modeled by distinguishing between private and public data (or protected and non-protected data). Second, and more importantly, copyright law excludes specific subject (e.g. ideas, methods of operation, facts), since they are regarded as raw materials needed for cultural expression. According to the US Supreme Court, ``originality'' is the ``sine qua non'' of copyright. \cite{Feist_Rural} Thus, only the original elements within copyrighted works are legally protected by copyright law. Unoriginal elements (e.g., ideas, facts) are never protected.
Privacy, in contrast, protects content and not expression, which in turn can be misaligned with the original objectives of copyright law.

This point cannot be overestimated. Copyright law not only allows subsequent authors to draw upon the unoriginal, and thus unprotected, elements of copyrighted works (unlike in privacy) but also encourages subsequent authors to do so \cite{litman1990public, elkin2015copyright}. Because copyright protection only applies to some elements within copyrighted works (i.e, expression) while deliberately excluding others (i.e., ideas) courts need to delineate the scope of legal protection when deciding copyright disputes. As a result, the scope of copyright protection varies not only among different works but also among different elements within a single work \cite{SAS_World}.

Third, in a stark distinction from privacy, copyright law also encourages  using the original (and thus protected) elements of copyrighted works in certain circumstances. These include de minimis quotations, transformative uses serving different purposes compared to the purpose of the original work (such as parodies), and other types of “fair uses” such as learning and research \cite{netanel2011making}. The fair use doctrine serves as a check on copyright, to ensuring it does not stifle the very creativity copyright law seeks to foster. Fair use is also considered one of the safety valves that allows copyright protection to coexist with freedom of expression  \cite{netanel2008copyright}.

For all these reasons, privacy notations are both over-inclusive and under-inclusive from a copyright perspective. They are over-inclusive because they withhold much more from subsequent authors than copyright law necessitates, consequently undermining the objectives of copyright law. At the same time, by focusing on content rather than original expression, privacy notations are also under-inclusive because they allow (in some cases) unlawful access to original copyrighted expression. This could happen, for example, if Alice’s model did not access input content B, but did access input content C that incorporated original expression deriving (lawfully or not) from input content B.

In this study we initiate a discussion about the challenges involved in providing a rigorous definition capturing the concept of copyright. We commence with a technical discussion, comparing different proposed notions of copyright (in particular, differential privacy and NAF) and examining their close connection to algorithmic stability. Subsequently, we argue that any approach following this line of reasoning encounters significant obstacles in modeling copyright as understood within the legal context. In more detail, we argue that algorithmic stability strategies fail to account for some principles of copyright law that intend to preserve copyright law's delicate balance. We identify several major gaps between algorithmic stability strategies and copyright doctrine. Accordingly, we argue, that if algorithmic stability techniques are adopted as a standard for copyright infringement, they may undermine the intended goals of copyright law. We further propose a different approach to using quantified measures in copyright disputes that could better reconcile copyright trade-offs.

\subsection{Related Work}

A growing number of researchers in recent years have explored how to address legal problems by applying computer science theories and methods . This literature seeks to narrow the gap between the vague and abstract concepts used by law and mathematical models, and  to offer more rigor, coherent, and scalable definitions for  issues such as privacy  \cite{dwork2018privacy}, fairness and discrimination. \cite{dwork2012fairness,kairouz2021advances}
In the context of generative models, \citet{carlini2023extracting} and \citet{haim2022reconstructing} have explored whether generative diffusion models memorize protected works that appeared in the models’ training set. Their approach indicates the mere possibility of unauthorized cpoying by GenAI models. However, as discussed, memorizing of the input content does not necessarily equate to  copyright infringement. To evaluate infringement we must consider other measurable metrics and quantified measures for copyright key limiting concepts. 

There is also active and thought-provoking discussion on how ML technologies are reshaping our understanding of copyright within the realm of law. 
\citet{asay2020independent} explores the question of whether AI system outputs should be subject to copyright protection. \citet{grimmelmann2015copyright, lemley2020fair} examine the implications of copyright law's notions of authorship and learning for literary machines. Our Focus, though, is on the legitimacy of using copyrighted materials by models that generate similar output content. 

The works of \citet{bousquet} and \citet{barak}w, hich rely on privacy/privacy-like notions, are the main focus of our work. An alternative approach taken by \citet{scheffler2022formalizing} proposes a framework to test the substantial similarity of a mode's output content by comparing Kolmogorov-Levin complexity with and without access to the copyrighted input content. However, one has to distinguish between protected expressions and non-protected ideas; this crucial challenge is overlooked by their approach. Another work by \citet{franceschelli2022deepcreativity} suggests using generative learning techniques to assess creativity. Such approaches may prove valuable, as we indicate in Section 4, but only if they are designed to align with copyright principles. Lastly, \citet{henderson2023foundation} seek to develop strategies to be applied to generative models to ensure they satisfy the same fair use standard as in human discretion. The application of this solution may not be possible, though, in cases where little to no open source or fair use data is readily available.

\section{Algorithmic stability as a surrogate for copyright}

In this section, we focus is introducing and discussing two notions of algorithmic stability: near-access-freeness (NAF) and differential privacy (DP); these two notions were specifically investigated in the realm of training methods aimed at safeguarding copyrighted data. 

NAF and DP adhere to a shared form of stability: they ensure that the resulting model, denoted as $q$, satisfies a safety condition with respect to each copyrighted data instance, denoted as $c$. This safety condition guarantees the existence of a ``safe model'', denoted by $q_c$, which does not infringe the copyright of data $c$, and importantly, $q$ exhibits sufficient similarity to $q_c$. Consequently, both NAF and DP guarantee that $p$ itself does not violate the copyright of the respective data instance $c$. 

%
%


Formally, we consider a standard setup of an unknown distribution $\mathcal{D}$, and a generative algorithm~$A$. The algorithm $A$, gets as an input a training set of i.i.d samples $S=\{z_1,\ldots, z_m\}\in Z^m \sim D^m$, and outputs a model $p^A_S=A(S)$, which is a distribution supported on $Z$. For simplicity, we will assume here that $Z$ is a discrete finite set, but of arbitrary size.
\citet{barak} consider a more general variant in which the output posterior is dependent on a ``prompt'' $x$, and $A$ outputs a mapping $p^{(A_S)}(\cdot | x)$ that may be regarded as a mapping from prompts to posteriors. For our purposes there is no loss in generality in assuming that $p$ is ``promptless'', and our results easily extend to the promptful case, by thinking of each prompt as inducing a different algorithm when we hard-code the prompt into the algortihm.



\paragraph{Differential Privacy} 

$A$ is said to be $(\alpha,\beta)$-differentially private \cite{dwork2006calibrating} if for every pair of input datasets $S,S'$ 
that differ on a single datapoint, we have that for every event $E$:
\begin{equation}\label{eq:privacy} 
\P(A(S) \in E) \le e^{\alpha} \P(A(S')\in E) + \beta \text{ and } \P(A(S') \in E) \le e^{\alpha} \P(A(S)\in E) + \beta
\end{equation}

The concept of privacy, viewed as a measure of copyright, can be explained as follows: Let's consider an event, denoted as $E$, which indicates that the generative model produced by $A$ violates the copyright of a protected content item $c$. The underlying assumption is that if the model has not been trained on $c$, the occurrence of event $E$ is highly improbable. Thus, we can compare the likelihood of the event $E$ when $c$ is present in the sample $S$ with the likelihood of $E$ when $c$ is not included in a neighboring sample $S'$ (which is otherwise identical to $S$). If $A$ satisfies the condition stated in equation \cref{eq:privacy}, then the likelihood of event $E$ remains extremely low, even if $c$ happened to be present once in its training set.


\paragraph{Near Access  Freeness} There are several shortcomings of the notion of differential privacy that have been identified. Some of these are reiterated in \cref{sec:law}. \citet{barak} proposed the notion of Near-Access Freeness (NAF) that relaxes differential privacy in several aspects. Formally, NAF (or more accurately NAF w.r.t safe function $\safe$ and $\Delta_{max}$ is defined as follows: First, we assume a mapping $\safe$ that assigns to each protected content $c$ a model $q_{c}$ which is considered safe in the sense that it does not breach the copyright of $c$.
The function $\safe$, for example, can assign $c$ to a model that was trained on a sample that does not contain $c$. 
Several $\safe$ functions have been suggested in \cite{barak}.

A model $p$ is considered $\alpha$-NAF if the following inequality holds simultaneously for every protected content $c$ and every $z$:
\begin{equation}\label{eq:naf}
p (z) \le e^{\alpha} q_{c}(z).
\end{equation}
The intuition behind NAF is very similar to the one behind DP, however there are key differences that can, in principle, help it circumvent the stringency of DP. 

\begin{enumerate}
    \item The first difference between NAF and DP is that the NAF framework allows more flexibility by picking the 'safe' function. Whereas DP is restricted to a safe model corresponding to training the learning algorithm on a neighboring sample excluding the content $c$. 
    \item A second difference is the fact that NAF is one sided (see \cref{eq:naf}), in contrast with DP which is symmetric (see \cref{eq:privacy}). Note that one-sidedness is indeed more aligned with the requirement of copyright which is non-symmetric. 
    \item NAF makes the distinction between content-safety and model-safety \cite{barak}. In more detail, the NAF notion requires that the output model is stable. This is in contrast with privacy that requires stability of the posterior distribution over the output models. In this sense the notion of NAF is more akin to \emph{prediction differential privacy} \cite{dwork2018privacy} then to differential privacy.
    \item Finally, NAF poses constraints on the model outputted by the learning algorithm (each constraint corresponds to a prespecified \emph{safe model}). This is in contrast with privacy which does not restrict the output model, but requires stability of the posterior distributions over output models. This distinction may seem minor but it can lead to peculiarities. For example, an algorithm that is completely oblivious to its training set and that always outputs original content can still violate the requirements of NAF. To see this, imagine that our learning rule outputs a model $q$ that always generates the same content $z$ which is completely original and not similar to any protected content $c$. However, depending on the safe models $q_c$ it can be the case that the model $q$ is not similar to any of them.
    
\end{enumerate}

These differences, potentially, allow NAF to circumvent some of the hurdles for using DP as a notion for copyright. For example, the one-sidedness seems sufficient for copyright and may allow models that are discarded via DP. Also, the distinction between model-safety and content-safety can, for example, allow models that may memorize completely the training set as long as a content they output does not provide a proof for such memorization. Next, the fact that NAF is defined by a set of constraints, and not a property of the learning algorithm, allows one to treat breaches of \cref{eq:naf} as soft ``flagging'' and not necessarily as hard constraints. This advantage is further discussed in \cref{sec:discussion}.
Finally, perhaps most distinguishable, is the possibility to use general safety functions that can capture copyright breaches more flexibly. We next discuss the implications of these refinements, and the question of model safety vs. content safety in NAF and in DP.

\paragraph{Model safety vs. Content safety}
Our first result is a parallel to Theorem 3.1 in \cite{barak} in the context of DP stability. Theorem 3.1 in \cite{barak} shows how to efficiently transform a given learning rule $A$ to a learning rule $B$ which is NAF-stable, provided that $A$ tends to output similar generative models when given inputs that are identically distributed. We state and prove a similar result by replacing NAF stability with DP stability, which demonstrates that the notion of DP can be relaxed , analogously to NAF, to require only content safety under proper assumptions:


Recall that the total variation distance between any two distributions is defined as:
$ \|q_1-q_2\| = \frac{1}{2}\sum |q_1(x)-q_2(x)|= \sup_{E} \left(q_1(E)-q_2(E)\right),$
\begin{proposition}\label{thm:main_up}
    Let $A$ be an algorithm mapping samples $S$ to models $q^A_S$ such that
    $ \E_{S_1,S_2}\left[\| q^A_{S_1}- q^A_{S_2}\|\right] \le \alpha,$
    where $S_1,S_2\sim D^m$ are two independent samples.
    Then, there exist an $(\epsilon,\delta)$ DP algorithm $B$ that receives a sample $S_B\sim D^{m_{priv}}$ such that if $m_{priv} = \tilde O\left(\frac{m}{\eta\epsilon}\log 1/\delta\right)$  and $S_A\sim D^m$ then:
    $  \E_{S_A,S_B}\left[\|\E[q^B_{S_B}]- q^A_{S_A}\|\right] \le \frac{2\alpha}{1+\alpha}+O(\eta).$
    Where the expectation within, is taken over the randomness of $B$.
\end{proposition}
The premise in the above theorem is identical to that in Theorem 3.1 in~\cite{barak} and captures the property that $A$ provides similar outputs on identically distributed inputs. The obtained algorithm $B$ is DP-stable and at the same time it has a similar functionality like $A$ in sense that its output model $q^B$ generates content $z$ which in expectation is distributed like contents generated by $q^A$.

\paragraph{Safety functions}
We now turn to a discussion on the potential behind the use of different safety functions. The crucial point  (which we discuss in great detail in \cref{sec:law} below) is that a satisfactory ``copyright definition'' \emph{must} allow algorithms to be highly influenced, even by their input content which is \emph{protected}. This reveals a stark contrast with algorithmic stability: it is easy to see that DP does not allow such influence. Indeed, the whole philosophy behind privacy is that a model is ``safe'' if it did not observe the private example (in particular was not influenced by it). 

This raises the question of whether the greater flexibility of the NAF model can provide better aligned notions of safety. In fact, if it is allowed to be influenced by protected data, one might even want to consider safe models that have \emph{intentionally} observed a certain content and derived out of it the derivatives that are not protected. 

The next result, though, shows that there is a \emph{no free lunch} phenomenon. For every protected content~$c$, we can either only consider safe models that observed $c$ and are influenced by it, or only safe models that \emph{never} observed it and were \emph{not} influenced by it.
In other words, if a protected content $c$ influenced its safe model $q_c$ then it must influence all safe models $q_{c'}$ for all protected contents $c'$.
We further elaborate on the implication of this result in \cref{sec:discussion}.

Below, $q_1$ and $q_2$ should be thought of as safe models, and $p$ as the model outputted by the NAF learning algorithm.
(So, in particular $p$ should satisfy \cref{eq:naf} w.r.t $q_1$ and $q_2$.)
This result complements Theorem~3.1 in \cite{barak} which shows that NAF can be satisfied in the sharded-safety setting when the two safe models are close in total-variation. The proof is left to \cref{prf:main_low}.
\begin{proposition}\label{thm:main_low}
Let $q_1$ and $q_2$ be two distributions such that
$ \|q_1-q_2\|\ge \alpha,$
then for any distribution $p$ we have that for some $z$:
$ p(z) \ge \frac{1}{2(1-\alpha)} \min\{q_{1}(z),q_{2}(z)\}.$
\end{proposition}

\section{The gap between algorithmic stability and copyright}\label{sec:law}
So far, we have provided a technical comparison between existing notions in the CS literature aimed at provable copyright protection. While the technical notion of privacy may seem closely related, as observed through NAF, there are differences. Accordingly, there is room for more refined definitions that could capture these essential differences. While algorithmic stability approaches hold promise in helping courts assess copyright infringement cases (an issue we further discuss in \cref{sec:discussion}), they cannot serve as a definitive test for copyright infringement. To see that, we next discuss the issue of copyright from a legal perspective. From this perspective, formal algorithmic stability approaches are both over inclusive and  under-inclusive. Consequently, we will organize this section based on these challenges.

\subsection{Over-inclusiveness }

Here we focus on a concern that algorithmic stability approaches may filter out lawful output content that does not infringe copyright in the input content. Because non-infringing output content is lawful, employing algorithmic stability approaches as filters to generative models may needlessly limit their production capabilities, and, thereby, undermine the ultimate objectives of copyright law.
Copyright law intends to foster the creation of original works of authorship by securing incentives to authors and, at the same time, ensuring the freedom of current and future authors to use and build upon existing works. The law derives from the U.S Constitutional authority: “To promote the Progress of Science and useful Arts, by securing for limited Times to Authors and Inventors the exclusive Right to their respective Writings and Discoveries.” 
\cite{US_constitution_1} 

However, promoting progress is often at odds with granting unlimited control over copyrighted materials. This is why copyright law sets fundamental limits on the rights granted to authors. Promoting progress is inconsistent with an unrestricted right to prevent every unauthorized use because creators and creative processes are embedded in cultural contexts. Creative processes often requires ongoing interactions with preexisting materials, whether through learning and research, engagement with prior art to generating new interpretations, or using a shared cultural language and applying existing styles to make works of authorship more comprehensible. Consequently, using copyrighted materials becomes a crucial input in any creative discourse \cite{cohen2012configuring, elkin1996cyberlaw}. 

For this reason, unlike the mandate of the algorithmic stability approaches, copyright law does not require output contents not to draw on input contents to be lawful. On the contrary, there are many cases where copyright law explicitly allows output contents to draw heavily on input contents without raising infringement concerns. In such cases, allowing input contents to impact output contents is not only something copyright law permits, but it is also something copyright law encourages. Doing so, as Jessica Litman put it, “is not parasitism; it is the essence of authorship.”  \cite{litman1990public}  

Copyright law allows output contents to substantially draw an input contents in three main cases, which we next explore: 
(1) When an input content is in the public domain, (2) When an input content is copyrighted but incorporates aspects excluded from copyright protection, and (3) When the use of the protected aspects of the input content is lawful.

\paragraph{When input content is in the public domain}
Input content may be unprotected because its copyright term has lapsed. Copyrights are limited in duration (though relatively long duration, which in most countries will last the life of the author plus seventy years). Once the copyright term expires, input content enters the public domain and can freely be used and impact output content without risking copyright infringement \cite{litman1990public}. 
Public domain materials may also contain anything that is not copyrightable, such as natural resources. For instance, if two photographers are taking  pictures of the same person, some similarity between those pictures is likely due to how this person looks, which is in the public domain. Other elements such as an original composition, or the choices made regarding lighting conditions and the exposure settings used in capturing the photograph, might be considered copyrighted expression. If the generative model only uses the former in the output content, it may not constitute an infringement.

\paragraph{When an input content incorporates unprotected aspects}
Input content with a valid copyright term enjoys “full” legal protection, but it too is limited in scope. As provided by the copyright statute, “[i]n no case does copyright protection for an original work of authorship extend to any idea, procedure, process, system, method of operation, concept, principle, or discovery, regardless of the form in which it is described, explained, illustrated, or embodied in such work.” \cite{17_U.S.C.}. By this principle, output content may substantially draw on input content without infringing copyright in the latter, as long as such taking is limited to the input’s content unprotected elements.
\begin{itemize}
\setlength\itemsep{-.2em}
    \item \textbf{Procedures, processes, systems and methods of operation}
    Copyright protection does not extend to “useful” or “functional” aspects of copyrighted works such as procedures, systems, and methods of operation. These aspects of an input content are freely accessible for an output content to draw upon. For example, in the seminal case of Baker vs. Selden, the Supreme Court allowed Baker to create a book covering an improved book-keeping system while drawing heavily on the charts, examples, and descriptions used in Selden’s book without infringing Selden’s copyright \cite{Baker_Selden}. As the court explained, these aspects that Baker took from Selden’s work are functional methods of operations and as such are not within the domain of copyright law.  Similarly, in Lotus v. Borland, the United States Court of Appeals for the First Circuit allowed Borland to copy Lotus’s menu command hierarchy for its spreadsheet program, Lotus 1-2-3. The court ruled that Lotus menu command hierarchy was not copyrightable because they form methods of operation \cite{lotus} - Consequently, if a generative model simply extracts procedures, processes, systems and methods from the training set it may not infringe copyright. 

    \item \textbf{Ideas}
    Copyright protection is limited to concrete “expressions” and does not cover abstract “ideas.” Thus, in Nicholas v. Universal, the United States Court of Appeals for the Second Circuit allowed Universal to incorporate many aspects of Anne Nichols’ play Abie’s Irish Rose, in their film The Cohens and Kellys \cite{nichols}.  The court explained that the narratives and characters that Universal used (“a quarrel between a Jewish and an Irish father, the marriage of their children, the birth of grandchildren and a reconciliation”), were “too generalized an abstraction from what she wrote. . . [and, as such]. . . only a part of her [unprotected] ‘ideas.’”  \cite{nichols}
    When a generative model simply extract ideas from copyrighted materials, rather than  replicating expressive content from their training data, it does not trigger  copyright infringement. 

    \item \textbf{Facts}
    Copyright protection also does not extend to facts. For example, in Nash v. CBS., the court ruled that CBS. could draw heavily from Jay Robert Nash’s books without infringing his copyright \cite{nash}. As the court explained, the hypotheses that Nash rose speculating the capture of the gangster John Dillinger and the evidence he gathered (such as the physical differences between Dillinger and the corpse, the planted fingerprints, and photographs of Dillinger and other gangsters in the 1930s) were all unprotected facts that Nash could not legally appropriate. Consequently, generative models which simply memorize facts do not infringe copyright law. 
\end{itemize}

\paragraph{When the use of the protected aspects of the input content was lawful}

Even when the protected elements of an input content (“expressions” rather than the “ideas”) are impacting an output content, such impact may be legally permissible. There are two main categories of lawful uses: de minimis copying and fair use.

\begin{itemize}
\setlength\itemsep{-.2em}
    \item \textbf{De minimis copying} 
    Copyright law allows de minimis copying of protected expression. I.e. copying of an insignificant amount that has no substantial impact on the rights of the copyright owner or their economic value. Similarly, “[w]ords and short phrases, such as names, titles, and slogans, are uncopyrightable.”\cite{works_not_protected}. However, de minimis copying of protected expression may be unlawful if it captures the heart of the work \cite{Harper_Row}.  E.g. phrases like “E.T. Phone Home.” \cite{universal_kamar}
    \item \textbf{Fair Use}
    Copyright law also allows copying of protected expression if it qualifies as fair use. The U.S fair use doctrine, as codified in § 107 of the U.S Copyright Act of 1976, is yet another legal standard to carve out an exception for an otherwise infringing use after weighing a set of four statutory factors. The four statutory factors are: (1) the purpose and character of the use, including whether such use is of a commercial nature or is for nonprofit educational purposes; (2) the nature of the copyrighted work; (3) the amount and substantiality of the portion used in relation to the copyrighted work as a whole; and (4) the effect of the use upon the potential market for or value of the copyrighted work \cite{17_U.S.C.}.

\end{itemize}
Importantly, the fair use claimant need not satisfy each factor for the use to qualify as fair use \cite{campell_acuff-rose}.  Nor are the four factors meant to set out some kind of mathematical equation whereby, if at least three factors favor or disfavor fair use, that determines the result \cite{netanel2011making}. Rather, the factors serve as guidelines for holistic, case-by-case decision.  In that vein, in its preamble paragraph, § 107 provides a list of several examples of the types of uses that can qualify as fair use. The examples, which include “criticism, comment, news reporting, teaching (including multiple copies for classroom use), scholarship, [and] research,”\cite{17_U.S.C.}  are often thought to be favored uses for qualifying for fair use. Importantly, however, the list of favored uses is not dispositive. Rather, fair use’s open-ended framework imposes no limits on the types of uses that courts may determine “fair” \cite{campell_acuff-rose}.  

When the factors strongly favor a finding of fair use, even output contents that are heavily impacted by copyrighted input contents may be excused from copyright infringement. For example, in Campbell v. Acuff-Rose, although the rap music group 2 Live Crew copied significant portions of lyrics and sound from Roy Orbison’s familiar rock ballad “Oh, Pretty Woman” \cite{campell_acuff-rose}. The Supreme Court denied liability in this case, based on the premise that the 2 Live Crew’s derivative work was considered a “parody” of Orbison’s original work, and, therefore, constituted fair use. Similarly, in The Authors Guild v. Google, the court defended Googles’ mass digitization of millions of copyrighted books to create a searchable online database as fair use, because it considered Google’s venture to be socially desirable \cite{Authors_Guild_Google} as explained by \citet{sag2018new}, concluding that the copying of expressive works for non-expressive purposes should not be counted as a copyright infringement. 
 
\subsection{Under-Inclusiveness}
Algorithmic stability approaches are under exclusive because they might fail to filter out unlawful output content that infringes copyright in the input content. As explained, algorithmic stability  approaches find infringement only when the output content heavily draws on input content. The law of copyright infringement, however, is not so narrow. Copyright law only requires that the output content heavily draw on the protected expression originating from an input content to find infringement. Such expression need not come from the input content itself; it may come from other sources including copies, derivatives or snippets of the original input content \cite{lee2023talkin}.

\begin{wrapfigure}{B}{0.3\textwidth}
\includegraphics[width=0.3\textwidth]{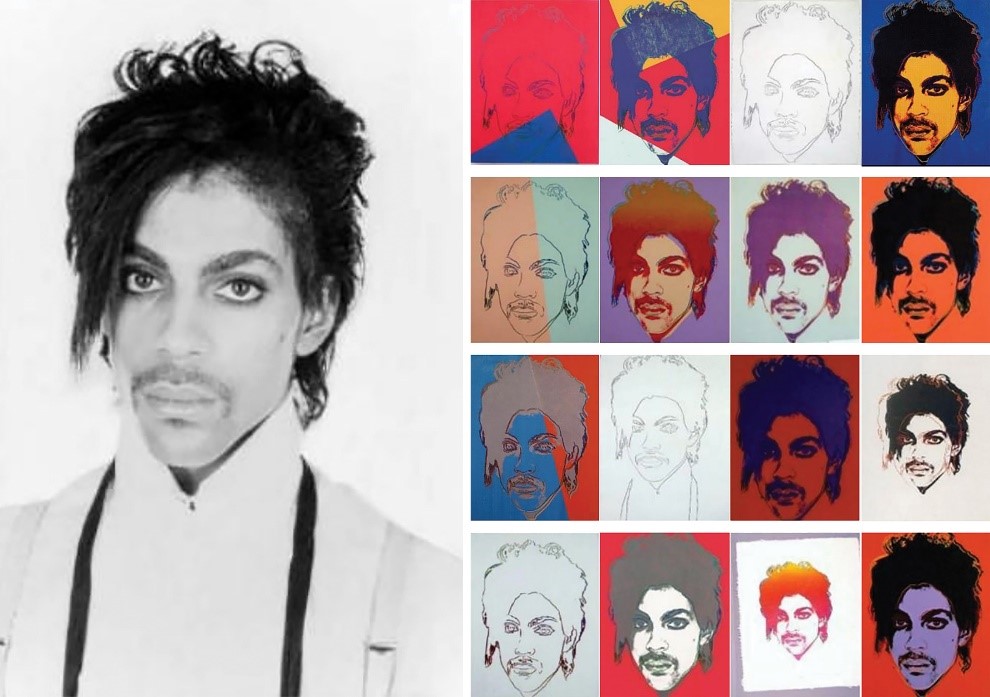}
    \caption{}\label{fig:prince}
\end{wrapfigure}

To illustrate this point, consider the fact pattern in the U.S Supreme Court case Warhol vs. Goldsmith \cite{Warhol_Goldsmith}.  In that case, the portrait photographer Lynn Goldsmith accused Andy Warhol of infringing copyrights in a photograph she took of the American singer Prince. Goldsmith authorized Warhol to use her photograph as an “artistic reference” for creating a single derivative illustration (see \cref{fig:prince}, bottom right most picture). Still, she did not approve nor imagine that Warhol had, in fact, made 16 different derivatives from the original photograph. Warhol’s collection of Prince portraits, also known as the Prince series , is depicted in \cref{fig:prince}, right side.

For our purposes, assume the Prince Series’ portraits served as input for a generative machine. Suppose the machine’s output content draws heavily on Goldsmith’s protected expression that is baked into the Prince Series’ portraits. In that case, the machine’s output content may infringe Goldsmith’s copyright in original photograph (\cref{fig:prince} , left side), even if the machine did not have access to Goldsmith’s original photograph. Moreover, this risk will not be eliminated even if the Supreme Court decided that the Prince Series' portraits themselves are non-infringing because they constitute fair use.

Simply put, copying from a derivative work—whether authorized by the copyright owner or not— may infringe copyright in the original work on which the derivative work is based. This situation is prevalent in copyright practice, especially in music.
In modern music copyright cases, plaintiffs usually show access to the original copyrighted work (musical composition) by showing access to a derivative work of that original work (sound recording).  Plaintiffs are not required to demonstrate that the defendants also had access to the original sheet music nor that they could actually read musical notes. 

Lastly, output content can also infringe copyright in input content by accessing parts or snippets of the input content even without accessing the input content in its entirety. This concern was raised recently in The Authors Guild v.\ Google, a case dealing with the legality of the Google Book Search Library Partner project \cite{Authors_Guild_Google}.  As part of this project, Google scanned and entered many copyrighted books into their searchable database but only provided “snippet views” of the scanned pages in search results to their users. The plaintiff in the case argued that Google facilitated copyright infringement by allowing users to aggregate different snippets and reconstruct infringing copies of their original works.  The court ended up dismissing this claim, but only because Google took affirmative steps to prevent such reconstruction by limiting the number of available snippets and by blacklisting certain pages. 

To sum up, there are numerous instances where copyright law permits (even encourages) an output content to draw on an input content. The more substantial unproteceted aspects of input content, and the more likely it is that using the input content’s protectable aspects is considered lawful, the more expansively can the output content draw upon the input content without fearing copyright infringement. At the same time, there are cases where copyright law outlaws an output even if it did not draw upon an input content, provided that it did draw on protected expression originating from that content. The more original the input content, and the more copies, derivatives, or snippets of that original content exist in the model datasets, the more likely the output content is to infringe copyrights in that input content. Therefore, any strategy for detecting or mitigating copyright infringement must account for these crucial copyright distinctions.

\section{Discussion}\label{sec:discussion}
Algorithmic stability approaches, when used to establish proof of copyright infringement are either too strict or too lenient from a legal perspective. Due to this misfit, applying algorithmic stability approaches as filters for generative models will likely to distort the delicate balance that copyright law aims to achieve between economic incentives and access to creative works.

The purpose of this article is to illuminate this misfit. This is not to say that algorithmic approaches in general and algorithmic stability approaches, in particular, have no value to the legal profession. Quite the opposite. Computer science methodologies significantly benefit the judicial table: the capability to process large volumes of information and assist policymakers in making more informed decisions.  Many areas in law involve applying murky “standards” as opposed to rigid “rules.” \cite{kaplow1992rules}. As discussed, copyright law extensively uses legal standards, such as idea/expression distinction, or fair use principles, to restrict the scope of protection accorded to copyrighted works. Consequently, copyright infringement cannot be boiled down to a binary computational test. 

The true value of computer science methodologies to the legal profession is not necessarily to convert murky standards into rigid rules (e.g., by constructing a definitive binary test for copyright infringement), but, instead, to make legal standards less murky. 
A rich body of scholarship explores the ills of vaguely-defined legal standards, especially in the context of intellectual property \cite{parchomovsky2009originality, benkler1999free, samuelson1996copyright, gibson2006risk, menell2013notice}
Algorithmic stability approaches, if applied with caution,  may introduce  new quantifiable methods for applying legal standards more clearly and predictably. Such methods could help measure vague legal concepts such as “fairness” “privacy,” and, in the copyright context—“originality”, and at the same time facilitate the ongoing development of legal and social norms \cite{hacohen2023copyright}. However, to ensure these methods are beneficial, it is vital to acknowledge the limitations of applying algorithmic stability approaches to copyright.  

\paragraph{Stability is not safe} The NAF framework, which allows a rich class of safety functions, has the potential to circumvent some of the challenges presented, but may still be limited and we now wish to discuss this in further details. RL is supported by an ERC Grant (FOG 

To utilize the NAF framework, the first basic question one needs to address is
\emph{
Given a protected content $c$ how should we choose the safe model $\safe(c)$?}
It seems natural to include models that are not heavily influenced by $c$ since otherwise this might allow copyright breaching. However, such choice of $\safe(c)$ leads to the discussed limitations encountered by algorithmic-stability approaches such as DP. It is true that some aspects, such as content safety vs. model safety, can be better aligned through the definition of NAF but also, as \cref{thm:main_up} shows, through variants of DP. Overall, there is room, then, to further investigate the different possible models for copyright, within such an approach, but we should take into account the limitations presented in \cref{sec:law}.

Perhaps a more exciting application of NAF, then, is to consider notions of safety that allow some influence by $c$. e.g.\ to enable generating parodies, fair-use, de minimis copying, etc. We consider then safety functions that now \emph{do} have access to $c$, and exploit this access to enable only allowed influence. Here we face a different challenge. Suppose that $q_c, q_{c'}$ are such a safe models for contents $c$ and $c'$ respectively. If $q_{c'}$ and $q_{c}$ are far away, then \cref{thm:main_up} shows that there is no hope to output a NAF model. But even if $q_{c}$ and $q_{c'}$ are not far away, but suppose that $q_{c'}$ ignores content $c$, then for any content $z$ that is influenced by $c$ we may assume that:
\[ q_{c}(z)\gg q_{c'}(z).\]
But, if $p$ is a NAF model, we must also have due to \cref{eq:naf} with respect to $c'$ and $z$:
\[ q_{c}(z) \gg p(z).\]
In other words, the NAF model censors permissible content $z$ even though it is safe. This happens because $z$ is an improbable event in model $q_{c'}$. Not because $z$ breaches copyright of $c'$ but because it is influenced by $c$, and content that is influenced by $c$ is discarded by safe models that had no access to $c$. 
It follows, then, that all safe models must treat protected content in a similar manner, and $q_{c'}$ must also be influenced by $c$ if we expect the NAF model to make any use of it. Hence, it is unclear if a more refined notion of $\safe$ may help circumvent the hurdles of applying the privacy approach for establishing a copyright infringement. This suggests, though, to perhaps consider a relaxed variant of NAF in which a content is discarded by a safe model only when certain links between the protected content and the generated content are established.



It seems, then, that an algorithmic approach that assists jurists in understanding such links between existing works of authorship, study their hidden interconnection, and quantify their originality holds a great promise. In other words, rather than constructing binary legal rules (e.g., aiming to devise a definitive test for copyright infringement), algorithmic stability approaches could facilitate new quantifiable methods for applying legal standards, such as measuring originality \cite{hacohen2023copyright}. From this perspective, originality is evaluated by the semantic distance between the elements of a measured expressive work and similar elements found in the corpus of the training content. The more salient the expressive elements within the larger corpus of pre-existing content, the less likely these elements are to be considered original by copyright law, and the more likely copyright law is to legitimize drawing upon them by the output content.

Research in this area is still in its infancy but holds outstanding potential for the copyright system \cite{shi2022selective, hartmann2023sok}. Algorithmic approaches that focus on the element level rather than the content level, and are applied not as binary tests for apprising infringement but as tools for measuring copyright originality may greatly empower the legal profession. As the extensive body of legal scholarship has long acknowledged, the originality standard in copyright law, along with many of its related doctrines for delineating scope (such as the “idea-expression dichotomy”), is inherently vague and uncertain \cite{litman1990public, lemley2009our, jones1990myth}. Such vagueness leads to inconsistent judicial precedent, deters permissible uses of copyrighted material, and undermines the goals of copyright law \cite{gibson2006risk, netanel2008copyright, litman2007billowing, samuelson1996copyright,vaidhyanathan2001copyrights}.

\paragraph{Acknowledgments}
We thank Bruria Friedman for research assistance.

\bibliographystyle{abbrvnat}
\bibliography{bibliography}

\appendix
\section{Proofs}
\subsection{Proof of \cref{thm:main_low}}\label{prf:main_low}
Suppose that \[\|q_{1}-q_{2}\|\ge \alpha.\] In particular there exists an event $E$ such that:

\[ q_{2}(E) \le q_{1}(E)- \alpha\le 1-\alpha.\]

Let $p$ be some distribution. We assume that $p(E)\ge 1/2$ (otherwise, replace $E$ with its complement and $q_1$ and $q_2$ replace roles). Thus, we have that:

\[p(E)\ge \frac{1}{2}\ge \frac{1}{2(1-\alpha)}q_{2}(E).\]
In particular, for some $z\in E$, the result follows.

\subsection{Proof of \cref{thm:main_up}}
The proof relies on a coupling Lemma, taken from \cite{angel2019pairwise}. Recall that, given a collection of distribution measures $Q$, a coupling can be thought of as a collection of random variables $X= (X_q)_{q\in Q}$, whose marginal distributions are given by $q$. I.e. $\P(X_q=x)=q(x)$:

\begin{lemma}[A special case of Thm 2 in \cite{angel2019pairwise}]\label{lem:coupling}
Let $Q$ be the collection of all posteriors over a finite domain $\X$\footnote{which are all absolutely continuous w.r.t the uniform distribution}. There exists a coupling such that for every $q,q' \in Q$:

\[\P(X_{q}\ne X_{q'}) \le  \frac{2\|q-q'\|}{1+\|q-q'\|}.\]
\end{lemma}
The second Lemma we rely on is a private heavy hitter mechanism, described as follows:

\begin{lemma}[\cite{korolova2009releasing, bun2016simultaneous}]\label{lem:histogram}
Let $Z$ be a finite data domain. For some
\[ k \ge \Omega\left(\frac{\log 1/\eta \beta \delta}{\eta \epsilon}\right),\]
there exists an $(\epsilon,\delta)$-DP algorithm $\hist$, such that with probability $(1-\beta)$ on an inputs $S = \{z_1,\ldots, z_k\}$ outputs a mapping $a\in [0,1]^Z $, such that, for every $z\in Z$, \[|a(z)-\textrm{freq}_{S}(z)|\le \eta.\]
In particular, if $\textrm{freq}_{S}(z)>0$, then $a(z)>0$.
\end{lemma}
Where we denote by $\textrm{freq}_S(z) = \frac{|i: z_i=z|}{|S|}$.

We next move on to prove the claim. Let $X$ be the coupling from \cref{lem:coupling}. Our private algorithm works as follows:
\begin{enumerate}
    \item First, we take $\beta = \eta$, and set
\[ k = \Omega \left(\frac{\log 1/\eta^2\delta}{\eta \epsilon}\right).\]
To be as in \cref{lem:histogram}.
    \item Divide $S$, the input sample, to $k$, disjoint datasets $S_1,\ldots, S_k$ of size $m$. 
    Each data set, via $A$, defines a model $q^A_{S_i}$. 
    \item Next, we define the random sample 
    \[S_{X} = \{ X_{q^A_{S_1}},X_{q^A_{S_2}},\ldots, X_{q^A_{S_k}}\} \in Z^K.\]
    \item Apply the mechanism in \cref{lem:histogram} and output $a\in [0,1]^Z$ such that, w.p. $1-\eta$, for all $z\in Z$:
    \[|a(z)-\textrm{freq}_{S_X}(z)|\le \eta.\]
    \item\label{step:3} Let $p$ be any arbitrary distribution such that for every $z\in Z$: \begin{equation}\label{eq:step3} |a(z)-p(z)|\le \eta\end{equation} (if no such distribution exists $p$ is any distribution). and output \[q_S^B=p.\]
\end{enumerate}
Notice that each sample $z_j$ affects only a single sub-sample $S_i$ and in turn only a single random variable $X_{q^A_{S_i}}$. The histogram function $a$ is then $(\epsilon,\delta)$-DP w.r.t to its input $S$. The output $p$, by processing is also private. We obtain, then, that the above algorithm is $(\epsilon,\delta)$-private.

We next set out to prove that $p=q_{S}^B$ is close in TV distance to $q^A_{S_A}$ in expectation. For ease of notation let us denote $X_i= X_{q^A_{S_i}}$. Notice that, with probability $(1-\eta)$, for every $z$:
\[ |a(z) - \textrm{freq}_{S_X}(z)|\le \eta,\]
in particular, there is a $p$ that satisfies the requirement in \cref{step:3} (i.e. $\textrm{freq}_{S_X}$ defines such a distribution) and \cref{eq:step3} is satisfied. We then have that for every $z$:

\begin{align*}   \left|p(z)-\frac{1}{k}\sum \mathbf{1}[X_i=z] \right| 
&\le \left|p(z)-a(z) \right|+ \left|a(z)-\frac{1}{k}\sum \mathbf{1}[X_i=z] \right|\\
&\le 2\eta \labelthis{eq:1} .\end{align*}

We now move on to bound the total variation between the model $\E[q^B_S]$ and $q_{S_A}$, where expectation is taken over the randomness of $B$. 

To show this, we will use the reverse inequality of the coupling Lemma, in particular if $(\hat X_B, \hat X_A)$ is a coupling of $q^B_S$ and $q^A_{S_A}$ (where $S$ and $S_A$ are now fixed), then: 

\begin{equation}\label{eq:couplingineq} \|\E[q^B_S]-q^A_{S_A}\|\le \P(\hat X_B\ne  \hat X_{A}).\end{equation}

Our coupling will work as follows, first we output $p= q^B_S$ and sample $\hat X_B \sim p$, and we let $\hat X_A= X_{q_{S_A}}$. This defines a coupling $(\hat X_B, \hat X_A)$.
Applying \cref{eq:1}, with $z= \hat X_A$, exploiting the fact that \cref{eq:1} holds with probability at least $1-\eta$:

\begin{align*}\P(\hat X_B \ne \hat X_A) 
&\le \frac{1}{k}\sum_{i=1}^k \P(X_i\ne X_{q_{S_A}}) + \eta\\
& \le 2\eta +\eta.
\end{align*}
And we have that:
\[ \P(\hat X_B \ne \hat X_{A}) \le  
\frac{1}{k}\sum_{i=1}^k \P(X_i\ne X_{q_{S_A}})+3\eta \le \frac{1}{k} \sum_{i=1}^k \frac{2\|q^A_{S_i}-q_{S_A}\| }{1+\|q^A_{S_i}-q_{S_A}\|} + 3\eta.\]
And,
\begin{align*}\E_{S_A,S}\|\E[q^B_{S}]-q_{S_A}\|&\le \E_{S_A,S}\frac{1}{k}\sum_{i=1}^k \left[\frac{2\|q^A_{S_i}-q_{S_A}\| }{1+\|q^A_{S_i}-q_{S_A}\|}\right] + 3\eta\\
&\le \E_{S_1,S_2\sim S}\left[\frac{2\|q^A_{S_1}-q_{S_2}\| }{1+\|q^A_{S_1}-q_{S_2}\|}\right] + 3\eta\\
& \le \left[\frac{2\E[\|q^A_{S_1}-q_{S_2}\|]}{1+\E[\|q^A_{S_1}-q_{S_2}]\|}\right] + 3\eta & \textrm{concavitiy of~}\frac{2x}{1+x}\\
& \le \left[\frac{2\alpha}{1+\alpha}\right] + 3\eta & \textrm{monotinicity~}\frac{2x}{1+x}
\end{align*}

\end{document}